\documentclass[conference]{IEEEtran}
\IEEEoverridecommandlockouts
\usepackage{cite}
\usepackage{amsmath,amssymb,amsfonts}
\usepackage{algorithmic}
\usepackage{graphicx}
\usepackage{textcomp}
\usepackage{amsthm}
\usepackage{multirow}
\usepackage{array}
\usepackage{subfig}

\usepackage{color}
\usepackage{xcolor}
\usepackage[colorinlistoftodos,prependcaption]{todonotes} 
\usepackage{hyperref} %adicionei

\def\BibTeX{{\rm B\kern-.05em{\sc i\kern-.025em b}\kern-.08em
    T\kern-.1667em\lower.7ex\hbox{E}\kern-.125emX}}
\begin{document}

%$\accVec{}{} = \left\{ \accData_{1}, \accData_{2}, ..., \accData_{n} \right\}$

\newcommand{\conjunto}[1]
{\ensuremath{\boldsymbol{\mathcal{\MakeUppercase#1}}}}

\newcommand{\accData}{\ensuremath{i}}

%------------------ Conjuntos -----------------------------

\newcommand{\inerciais}{\ensuremath{\conjunto{I}}}
\newcommand{\accVec}[2]{\ensuremath{\inerciais_{#1}^{\,#2}}}

\newcommand{\high}{\ensuremath{\conjunto{H}}}
\newcommand{\HIGH}[2]{\ensuremath{\high_{#1}^{\,#2}}}

\newcommand{\low}{\ensuremath{\conjunto{L}}}
\newcommand{\LOW}[2]{\ensuremath{\low_{#1}^{\,#2}}}

\newcommand{\collor}{\ensuremath{\conjunto{C}}}
\newcommand{\COLLOR}[2]{\ensuremath{\collor_{#1}^{\,#2}}}

\newcommand{\noise}{\ensuremath{\conjunto{N}}}
\newcommand{\NOISE}[2]{\ensuremath{\noise_{#1}^{\,#2}}}

\newcommand{\bright}{\ensuremath{\conjunto{B}}}
\newcommand{\BRIGHT}[2]{\ensuremath{\bright_{#1}^{\,#2}}}

\newcommand{\unseen}{\ensuremath{\conjunto{U}}}
\newcommand{\UNSEEN}[2]{\ensuremath{\unseen_{#1}^{\,#2}}}

\newcommand{\fusion}{\ensuremath{\conjunto{F}}}
\newcommand{\FUSION}[2]{\ensuremath{\fusion_{#1}^{\,#2}}}

%------------------ Model -----------------------------

\newcommand{\model}[1]{\ensuremath{{\mathcal{\MakeUppercase#1}}}}
\newcommand{\DIFFUSION}{\ensuremath{\model{D}}}

%\title{Conference Paper Title*\\
\title{UDBE: Unsupervised Diffusion-based Brightness Enhancement in Underwater Images\\

%\title{UBRDM: Unsupervised Brightness Restoration in Underwater Images from Diffusion Models\\
%{\footnotesize \textsuperscript{*}Note: Sub-titles are not captured for https://ieeexplore.ieee.org  and should not be used}
}

\author{\IEEEauthorblockN{1\textsuperscript{st} Tatiana Taís Schein}
%Centro de Ciências Computacionais (C3). Universidade Federal do Rio Grande, Rio Grande -  RS, Brasil.
\IEEEauthorblockA{\textit{Centro de Ciências Computacionais} \\
\textit{Universidade Federal do Rio Grande}\\
Rio Grande -  RS, Brazil \\
tatischein@furg.br}
\and
\IEEEauthorblockN{2\textsuperscript{nd} Gustavo Pereira de Almeira}
\IEEEauthorblockA{\textit{Centro de Ciências Computacionais} \\
\textit{Universidade Federal do Rio Grande}\\
Rio Grande -  RS, Brazil \\
gustavo.pereira.furg@furg.br}
\and
\IEEEauthorblockN{3\textsuperscript{rd} Stephanie Loi Brião}
\IEEEauthorblockA{\textit{Centro de Ciências Computacionais} \\
\textit{Universidade Federal do Rio Grande}\\
Rio Grande -  RS, Brazil \\
stephanie.loi@furg.br }
\and
\IEEEauthorblockN{4\textsuperscript{th} Rodrigo Andrade de Bem}
\IEEEauthorblockA{\textit{Centro de Ciências Computacionais} \\
\textit{Universidade Federal do Rio Grande}\\
Rio Grande -  RS, Brazil \\
rodrigobem@furg.br}
\and
\IEEEauthorblockN{5\textsuperscript{th} Felipe Gomes de Oliveira}
\IEEEauthorblockA{\textit{Instituto de Ciências Exatas e Tecnologia} \\
\textit{Universidade Federal do Amazonas}\\
Itacoatiara -  AM, Brazil \\
felipeoliveira@ufam.edu.br}
\and
\IEEEauthorblockN{6\textsuperscript{th} Paulo L. J. Drews-Jr}
\IEEEauthorblockA{\textit{Centro de Ciências Computacionais}  \\
\textit{Universidade Federal do Rio Grande}\\
Rio Grande -  RS, Brazil \\
paulodrews@furg.br}
}

\maketitle

%=================================================================

\begin{abstract}

Activities in underwater environments are paramount in several scenarios, which drives the continuous development of underwater image enhancement techniques. A major challenge in this domain is the depth at which images are captured, with increasing depth resulting in a darker environment. Most existing methods for underwater image enhancement focus on noise removal and color adjustment, with few works dedicated to brightness enhancement. This work introduces a novel unsupervised learning approach to underwater image enhancement using a diffusion model. Our method, called UDBE, is based on conditional diffusion to maintain the brightness details of the unpaired input images. The input image is combined with a color map and a Signal-Noise Relation map (SNR) to ensure stable training and prevent color distortion in the output images. 
%The metrics and loss functions were carefully designed to tackle the addressed problem. 
%
The results demonstrate that our approach achieves an impressive accuracy rate in the datasets UIEB, SUIM and RUIE, well-established underwater image benchmarks. Additionally, the experiments validate the robustness of our approach, regarding the image quality metrics PSNR, SSIM, UIQM, and UISM, indicating the good performance of the brightness enhancement process. The source code is available \href{https://github.com/gusanagy/UDBE}{here}.

\end{abstract}


%=================================================================

\section{Introduction}
\label{sec:intro}

%\felipe{Discutir sobre dificuldade de ter imagens de referencia para ambientes subaquáticos. Explicar que é não supervisionado pq não utiliza imagens pareadas no treino.}

The underwater environment is influenced by several factors that negatively impact visibility and consequently affect the quality of underwater images and videos. 
The presence of suspended sediments in the water, such as sand, plankton, algae, and other organic matter blocks the passage of light, making the environment darker and hazier. They also reflect incoming light, creating an uneven haze that distorts the colors of submerged objects.
Another critical factor is the density of water, which is approximately 800 times greater than that of air. This density significantly attenuates visible light as depth increases, making visual clarity even more difficult. Thus, these combined factors make the visual exploration of the underwater environment a considerable challenge~\cite{fayaz2020underwater}.

Underwater low light conditions distort colors, resulting in bright spots in the center of images and darker edges. Artificial lighting, which is susceptible to this distortion, involves heavy and costly equipment that requires constant power, either from batteries or surface connections. Enhancing underwater images is a deeply explored field, with various methods investigated to achieve this goal. However, there is still limited research on low-light underwater images. Especially due to limited underwater datasets for research, making studies even more challenging when paired images are not available~\cite{abbas2019}.

To tackle this problem, different algorithms and methods have been proposed to mitigate the impacts of low light scattering and absorption. In this scenario diffusion models arise as a potential solution to attenuate the negative effects caused by limited lighting conditions. Diffusion models are deep generative models composed of two main stages: $i)$ the direct diffusion, including gradual Gaussian noise into the raw image; and $ii)$ the reverse diffusion, which learns to remove the noise and to generate a clear output image~\cite{ho2020denoising}.

In this paper, we propose a novel unsupervised learning method called UDBE, based on the Denoising Diffusion Probabilistic Model (DDPM)~\cite{ho2020denoising}, for enhancing underwater images. Our approach enhances the visibility and quality of subaquatic images, supporting visual exploration of the underwater environment, even without using reference images in the learning process (paired images). To achieve this, we use the U-Net network~\cite{ronneberger2015u}, trained with conditional and unconditional data. The conditional data are provided by a brightness module incorporated in every U-Net layer. 
Thereby, by combining low-lighting images, SNR maps, and raw images, the illumination conditions are compensated, providing visible and good-quality underwater images. The network was validated using PSNR, SSIM, UIQM, and UISM metrics.

The two main contributions are summarized as follows:

\begin{itemize}

  \item We propose an unsupervised learning method for lighting and turbidity enhancement in underwater images, overcoming the natural lack of reference images in underwater image datasets. We show that our method does not need manual human annotation to outgrow the state-of-the-art techniques in different datasets.
  \item We introduce a brightness control module that retains the brightness features of the original images. To our knowledge, UDBE is among the pioneering approaches using diffusion models for brightness enhancement in underwater images.
\end{itemize}

%=================================================================

\section{Related Work\label{sec:related}}

A wide range of applications for underwater image enhancement can be found, including navigation, offshore engineering, and monitoring \cite{Martinho2024}. Related techniques can be divided into three categories: physics-based, traditional, and deep learning-based methods. The following subsections introduce the most related works found in the literature.

\subsection{Physics-based Underwater Image Enhancement}

Recently, researchers have proposed different estimation methods for recovering degraded underwater images. The essence of physical model-based methods is to establish an underwater image formation model with prior knowledge, such as Dark Channel Prior (DCP)~\cite{he2010single} that removes haze. However, the concentration and dispersion of light in the underwater environment block some adaptations of the method. Peng et al.~\cite{peng2017underwater} developed an improved DCP method that uses the degree of image blur and the difference in light absorption to estimate the transmission map of underwater scenes. Peng et al.~\cite{peng2018generalization} created a Generalized Dark Channel Prior (GDCP) for image restoration, which incorporates adaptive kernel correction into an image formation model.

Although physics-based models can be effective in certain underwater environments, their limitations are evident. Inaccuracy in estimating the degradation process, especially in dynamic underwater scenarios, and lack of consideration for the human visual system are significant challenges.

\subsection{Traditional Underwater Image Enhancement}

Different traditional approaches have tackled the underwater image enhancement problem through intensity transformation techniques.
Huang et al.~\cite{huang2018shallow} proposed a Relative Global Histogram Stretching (RGHS) algorithm, mainly based on histogram equalization and stretching. 
Hassan et al.~\cite{hassan2021retinex} have employed the Retinex theory, a classic method for image restoration. These authors enhanced the underwater images using CLAHE, and then the Retinex theory is applied to restore the distorted colors.
Zhuang et al.~\cite{zhuang2021bayesian} developed a Bayesian Retinex algorithm to enhance single underwater images with prior multi-order gradients of reflectance and illumination. Chen et al.~\cite{chen2023enhancement} present a hybrid underwater enhancement method that solves an inverse problem with the new Retinex transmission map estimation and adaptive color correction. Martinho et al.~\cite{UIEFIT} proposed an approach combining intensity transforming techniques. Color, gamma, contrast, and brightness intensity corrections were combined to compensate for the underwater image degradation.

Traditional underwater image enhancement methods apply individual transformation techniques to improve the quality of underwater images, or even combine different transformation techniques. However, the aforementioned approaches are limited to fixed settings and parameters, making them ineffective for different and dynamic underwater environments.

\subsection{Deep Learning-based Underwater Image Enhancement}

Deep learning-based techniques have approached the enhancement of underwater images through a wide range of strategies.  
Wang et al.~\cite{wang2021uiec} presents the UIECˆ2-Net network using two color spaces for performing subaquatic image enhancement. This method introduces a CNN architecture to learn visual features from distinct color spaces, intending to be more sensitive to luminance and saturation issues. Martinho et al.~\cite{Martinho2024} introduces an adaptive learning model that estimates optimal (or near-optimal) parameters for intensity transformation techniques. The method applies a CNN architecture for the parameter estimation.

Another learning-based field considers the employment of attention mechanisms. Li et al.~\cite{li2021underwater} proposed Ucolor, which gathers features from three color spaces to enrich representations. Using an attention module, Ucolor integrates and highlights distinct features across a network of encoders. Additionally, it employs a network of transmission guide decoders, using the reversed-medium transmission (RMT) map as input, to improve response in degraded areas. These methods also consider the depth map. Wang et al.~\cite{wang2020gan} used class condition attention, in which an underwater image is classified first and then the class label guides the generation of the enhanced images. Fu et al.~\cite{fu2021underwater} uses residual dual attention to extract and enhance features with non-local and channel attention. Due to variable lighting and the sea depth, these approaches fall short in image restoration.

% \textcolor{blue}{Saleh et al. \cite{saleh2023distribuicao} realizaram o aprimoramento de imagens subaquáticas não supervisionado. O modelo, chamado UDnet, combina um U-Net como extrator de características com PAdaIN para codificar a incerteza, eliminando a necessidade de anotações manuais. }
%
Saleh et al.~\cite{saleh2023distribuicao} performed unsupervised underwater image enhancement. The model, called UDnet, combines a U-Net as a feature extractor with PAdaIN to encode uncertainty, eliminating the need for manual annotations.
%
% \textcolor{blue}{Mello et al. \cite{Claudio} apresentaram uma metodologia de aprimoramento de imagens subaquáticas auto-supervisionada, sem a necessidade de dados pareados. A proposta envolve a estimação da degradação presente nas imagens subaquáticas e a reconstrução dessas imagens por meio de um autoencoder. A imagem de saída do autoencoder é degradada usando a informação de degradação estimada e substituída na função de perda durante o treinamento. Isso engana a rede neural, que aprende a compensar a degradação adicional, resultando em uma versão aprimorada da imagem de entrada. Além disso, um módulo de atenção é incorporado para reduzir áreas de alta intensidade geradas por desequilíbrios nos canais de cor e regiões de pixels intensos.}
%
Mello et al.~\cite{Claudio} proposed a self-supervised underwater image enhancement method that estimates image degradation and uses an autoencoder for reconstruction. The network learns to compensate for degradation by incorporating it into the loss function. An attention module mitigates high-intensity areas caused by color imbalances.

Recently, diffusion models have emerged for enhancing underwater images. Chen et al.~\cite{zhao2023wavelet} proposed WF-Diff, which uses frequency domain information features and diffusion models. This framework includes WFI2-net, which enhances the frequency information in the wavelet space, and FRDAM, which refines the high and low-frequency information of the enhanced images. Dazhao et al.~\cite{du2023uiedp} presented UIEDP for underwater image enhancement, a network that samples the posterior distribution of sharp images conditioned on degraded images. Lu et al.~\cite{lu2023underwater} introduced UW-DDPM, an enhancement framework based on the Diffusion Model. This method uses two U-Net networks to effectively remove noise from underwater images, significantly improving image quality and reducing artifacts compared to traditional techniques. Tang et al.~\cite{tang2023underwater} developed a probabilistic conditional denoising diffusion model to generate enhanced underwater images. Their approach uses underwater images and Gaussian noise as inputs, applying a diffusion process to improve image quality. The method incorporates a transformer-based denoising network and skip sampling strategies to enhance efficiency and performance.

Our work proposes a new unsupervised approach for enhancing brightness in underwater images using Diffusion Models, called UDBE. To the best of our knowledge, this is the first time diffusion models have been employed to enhance brightness in underwater images. The proposed method also performs conditional enhancement, that is, it maintains the brightness features of the original images. Another positive point is that paired data is not required.

%=================================================================

%\input{Sections/Problem}

%=================================================================

\section{Methodology\label{sec:methodology}}

\begin{figure*}[ht]
\centering
\includegraphics[width=0.99\linewidth]{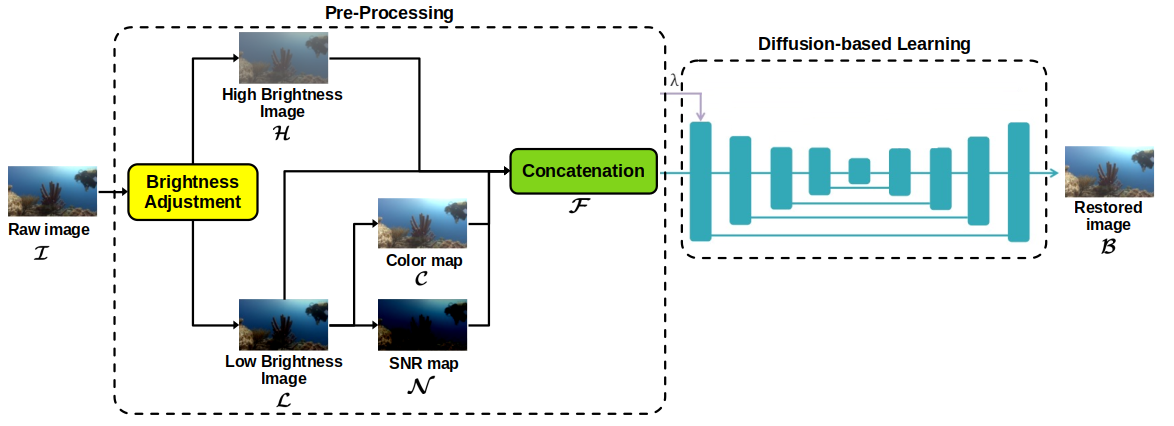}
\caption{Overview of the proposed methodology for brightness restoration in underwater images. The proposed approach is composed of two main steps: $i)$ a Pre-Processing stage, to extract and represent features for the learning process; and $ii)$ Diffusion-based Learning, to teach the model to compensate for the brightness degradation in underwater images.} 

\label{Metodologia}
\end{figure*}

We introduce the UDBE, an unsupervised diffusion-based approach for improving underwater images. Our approach consists of two primary steps: $i)$ Image Pre-Processing, which extracts and represents crucial information for the learning process in the Diffusion model; and $ii)$ Diffusion-based Learning Process, which through the diffusion model learns to compensate for brightness variation in underwater images. Figure~\ref{Metodologia} presents an overview of the proposed approach.

\subsection{Problem Statement}

\textbf{[Brightness Enhancement in Underwater
Images]}
Let $\accVec{}{} = \left\{ \accData_{1}, \accData_{2}, ..., \accData_{n} \right\}$ be a set of raw underwater images. Let $\HIGH{}{}$ be a set of high-brightness underwater images and $\LOW{}{}$ be a set of low-brightness underwater images, computed from $\accVec{}{}$. Also, let $\COLLOR{}{}$ be a series of color maps and $\NOISE{}{}$ be a series of noise representation maps, computed from $\LOW{}{}$. 
Let $\DIFFUSION$ be an unsupervised Diffusion Model, trained from $\HIGH{}{}$, $\LOW{}{}$ and $\FUSION{}{}$, the fusion between $\COLLOR{}{}$ and $\NOISE{}{}$.
Our main goal is to predict brightness-enhanced underwater images $\BRIGHT{}{}$, using the unsupervised Diffusion Model $\DIFFUSION$, from unseen raw underwater images $\UNSEEN{}{}$.

\subsection{Image Pre-Processing}

In this step, a series of raw underwater images ($\accVec{}{}$) in Red, Green, and Blue ($RGB$) color space is processed to extract and represent important features for the unsupervised learning process. To achieve this, different operations are carried out, such as Brightness Adjustment, Color Map Generation, Noise Representation Map Generation, and Combining Color and Noise Representations, whose details will be presented as follows.

\subsubsection{Brightness Adjustment} To teach the diffusion model ($\DIFFUSION$) to compensate for brightness degradation in images acquired in underwater environments, high and low-brightness underwater images are generated. Thereby, the instances of different brightness conditions provide important features regarding the brightness variation improving the learning stage. 

For this, the high-brightness underwater images ($\HIGH{}{}$) are obtained through the computation of random numbers representing bright values to be added to the raw images. The random numbers for the high-brightness images are comprised of the range [50, 100]. For the low-brightness underwater images ($\LOW{}{}$), random numbers representing bright values are also computed to be subtracted from the raw images. The random numbers for the low-brightness images are comprised of the range [-50, -100].

\subsubsection{Color Map Generation}

In addition to brightness, another important feature related to the quality of underwater images is color. In this step, the color features are computed from $\LOW{}{}$ images and depicted as color maps ($\COLLOR{}{}$). The color maps reduce color distortion by normalizing the range of the three channels in the input images \cite{yin2023cle}. Equation~\eqref{color} is given as follows:

\begin{equation}
Color(l)=\frac{l}{l_{\textit{max}}}=\left[\frac{l_{\textit{r}}}{l_{\textit{r}_{max}}},\frac{l_{\textit{g}}}{l_{\textit{g}_{max}}},\frac{l_{\textit{b}}}{l_{\textit{b}_{max}}}\right],
\label{color}
\end{equation}

\noindent where the low-brightness underwater image $(l)$ is decomposed into three channels, red ($l_r$), green ($l_g$), and blue ($l_b$). Meanwhile, $l_{max}$ represents the maximum value in the $l$ image.

\subsubsection{Noise Representation Generation}

A relevant aspect, also related to the quality of underwater images, is noise. In this stage the noise feature is represented as Signal-to-Noise Ratio (SNR) maps ($\NOISE{}{}$), computed from $\LOW{}{}$ images.
The SNR maps are used to visually highlight the areas where the signal-to-noise ratio is lowest. It is calculated according to the equation~\eqref{SNR} as follows,

\begin{equation}
Noise(l)=\frac{G(l)}{|l-G(l)+\epsilon|}.
\label{SNR}
\end{equation}

\noindent We use the parameter $\epsilon $ to guarantee numerical stability and the filter $G$ as a Gaussian blur. We define the high-frequency component of the image as noise and directly calculate the relationship between the original image and this noise.

\subsubsection{Combining Color and Noise Representations}

Finally, after the processing of brightness, color, and noise features in the previous steps, the color maps ($\COLLOR{}{}$) and SNR maps ($\NOISE{}{}$) are combined through a concatenation operation, as defined in the equation below: 

\begin{equation}
\FUSION{}{} = Concatenate[ \COLLOR{}{}, \NOISE{}{} ].
\label{concat}
\end{equation}

Our motivation for the color and noise maps concatenation is due to the loss of features during the diffusion model learning, thereby, the concatenation provides meaningful features to be highlighted in the learning process.
In this sense, the inputs for the diffusion model are the high-brightness underwater images ($\HIGH{}{}$), the low-brightness underwater images ($\LOW{}{}$), and the concatenation of color maps and SNR maps ($\FUSION{}{}$). 

\subsection{Diffusion-based Learning Process}

In this step, a CLE Diffusion-based model is used to enhance the brightness in degraded underwater images. It is important to highlight that the CLE Diffusion model \cite{yin2023cle} is not employed in the underwater domain, thereby the proposed diffusion-based learning model addresses the enhancement of lighting in subaquatic images. Additionally, we designed domain-specific loss functions tailored to our needs. It is worth mentioning that the proposed unsupervised learning process is based on the lack of reference images in the dataset, for the learning stage. For this, the low-brightness underwater images created in the Image Pre-Processing step are used as input data and the high-brightness underwater images are used as reference data.
%\felipe{Verificar essa frase com o Gustavo. Na análise quantitativa é dito que a low-brightness image é o input e a raw image é a referencia. Está correto?}

\subsubsection{Diffusion Model}

A Diffusion model is a probabilistic generative model that uses an iterative approach to create data. This model consists of a forward diffusion process that introduces noise to clean images and a reverse diffusion process that restores the clean images.

In this stage, the diffusion process is a Markov chain, which adds noise to the input data $x_{0}$ at each time interval $t$. Equation~\eqref{difdireta} describes the forward diffusion process using the DDPM model \cite{ho2020denoising}.

\begin{equation}
q(x_{t}|x_{t-1}) = N(x_{t}; \sqrt{1-\beta _{t}}x_{t-1}, \beta _{t}I),
\label{difdireta}
\end{equation}

\noindent where $N$ represents the Gaussian distribution, which is defined by the mean $\sqrt{1-\beta _{t}}x_{t-1}$ and the variance $\beta _{t} I$. The sequence $\beta_{1},...,\beta_{T}$ is a fixed noise variance schedule, which must converge to pure random Gaussian noise $N(0,1)$. Defining $\alpha_{t}=1-\beta_{t}$ and $\bar{\alpha}_{t}=\prod _{j=1}^{t} \alpha_{j}$, we can sample $x_{t}$ at any time step, as described in equation~\eqref{amostra}:

\begin{equation}
x_{t}= \sqrt{\bar{\alpha}_{t}}x_{0}+\sqrt{1-\bar{\alpha}_{t}}\epsilon,
\label{amostra}
\end{equation}

\noindent where $\epsilon$ represents Gaussian noise. The reverse diffusion process using DDPM model can be defined by the equation~\eqref{reversa}:

\begin{equation}
p_\theta (x_{t-1}|x_{t})= N(x_{t-1};\mu _{\theta }(x_{t}, t), \Sigma_{\theta}(x_{t},t)),
\label{reversa}
\end{equation}

\noindent where $\mu _{\theta }(x_{t}, t)$ denotes the mean and $\Sigma_{\theta}(x_{t},t)$ the variance of the distribution. It is worth mentioning that in the DDPM model, only the mean is learned and the variance is fixed~\cite{ho2020denoising}. Equation~\eqref{mean} defines the mean $\mu_{\theta}(x_{t}, t)$ as:

\begin{equation}
\mu _{\theta}(x_{t},t)=\frac{1}{\sqrt{\alpha_{t} }}\left(x_{t}- \frac{\beta_{t}}{\sqrt{1-\bar{\alpha }_{t}}} \epsilon _{\theta } (x_{t},t)\right),
\label{mean}
\end{equation}

\noindent where $\epsilon _{\theta }$ is obtained with a U-Net model~\cite{ronneberger2015u} to estimate the noise component of an image.
In addition to the diffusion model, a brightness control module assumes the input of a brightness level ($\lambda$), that improves the quality of features extracted by the U-Net model. A FiLM layer \cite{perez2018film}, is added to learn an affine transformation incorporating the light features to be applied during the unsupervised diffusion learning process employed in this work.

During inference, the noise is gradually reduced until a clear image is obtained. Additionally, to speed up the sampling process, the DDIM (Denoising Diffusion Implicit Models) sampler is used, following~\cite{song2020denoising} and defined as in equation~\eqref{DDIM}:

{\footnotesize
\begin{equation}
 x_{t-1}=\sqrt{\bar{\alpha}_{t-1}}\left( \frac{x_{t}-(\sqrt{1-\bar{\alpha}_{t}})\epsilon_{\theta}(x_{t},t)}{\sqrt{\bar{\alpha}_t}}\right)+(\sqrt{1-\bar{\alpha}_{t-1}})\epsilon_{\theta }(x_{t},t).
\label{DDIM}
\end{equation}
}

Model training takes place through the optimization of the negative log-likelihood loss function~\cite{ho2020denoising}, given by equation~\eqref{optimization} in a simplified form:

{\small
\begin{equation}
M_{simple}(\theta)= E_{x_{0}, t, \epsilon} \left[\left \| \epsilon -\epsilon_{\theta } (\sqrt{\bar{\alpha }_{t}}x_{0}+ \sqrt{1-{\bar{\alpha }_{t}}}\epsilon, t)\right\|^{2}\right].
\label{optimization}
\end{equation}
}

\subsubsection{Loss Functions}

A combination of different loss functions to address the brightness enhancement in underwater images is proposed for the unsupervised diffusion-based learning stage, as shown in equation~\eqref{loss}. 

\small{
\begin{equation}
L= \gamma_1 L_{LPIPS} + \gamma_2 L_{SSIM} + \gamma_3 L_{MSE} + \gamma_4 L_{brightness}+\gamma_5 L_{color}
\label{loss}
\end{equation}
}

\noindent where for $i=1,2,3,4,5$ the weights $\gamma_i$ are values $\gamma_1=30$, $\gamma_2=2.83$, $\gamma_3=1$, $\gamma_4=20$ and $\gamma_5=100$. The aforementioned weights were empirically defined in order to improve the brightness enhancement of underwater images, regarding the contribution of every component for the combined loss function.

%\felipe{Verificar essa equação. Está certa?}

The applied loss functions tackle distinct features, improving the overall learning. The employed loss functions are presented below:\\
\textbf{Perceptual Loss:} The LPIPS \cite{zhang2018unreasonable} perceptual loss function was used, which extracts high-level features using a pre-trained AlexNet network, trained on the ImageNet dataset. Additionally, it is calculated the distance between the enhanced features and the original features. This can be computed using the equation below:

\begin{equation}
L_{LPIPS}= \sum_{k} \frac{1}{H_{k}W_{k}}|\phi _{Alex}^{k}(\BRIGHT{}{})-\phi _{Alex}^{k}(\accVec{}{})|_{2}
\label{perceptual}
\end{equation}

\noindent where $\phi _{Alex}^{k}$ represents the feature maps extracted from the $k-th$ layer of the AlexNet model, and $H_k$, $W_k$ are the height and width of the feature map in layer $k$, respectively. The LPIPS loss function is used due to its ability to enhance the model for restoring high-frequency information \cite{zhang2018unreasonable}.

\textbf{SSIM Loss:} It calculates the structural similarity between the real image and the resulting image, providing statistics on overall contrast and luminance consistency \cite{wang2004image}. It can be computed using the equation \eqref{SSIM}:

\begin{equation}
L_{SSIM}=\frac{(2\mu _{\accVec{}{}}\mu _{\BRIGHT{}{}}+c_{1})(2\theta _{\accVec{}{}\BRIGHT{}{}}+c_{2})}{(\mu^{2} _{\accVec{}{}}+\mu_{\BRIGHT{}{}}^{2}+c_{1})(\theta^{2} _{\accVec{}{}}+\theta_{\BRIGHT{}{}}^{2}+c_{2})},
\label{SSIM}
\end{equation}

\noindent where $ \mu_{\accVec{}{}}$ and $ \mu_{\BRIGHT{}{}}$ are the mean pixel values, $\theta_{\accVec{}{}}$ and $\theta_{\BRIGHT{}{}}$ are the variances, $\theta_{\accVec{}{}\BRIGHT{}{}}$ is the covariance, and $c_{1}$ and $c_{2}$ are constants for numerical stability.

\textbf{Angular Color Loss:} It ensures that the colors of the enhanced images $\BRIGHT{}{}$ match the ground truth $\accVec{}{}$ \cite{wang2019underexposed}. The color loss can be expressed as the equation below:

\begin{equation}
%L_{\text{color}} =  \sum_{d} < ({\BRIGHT{}{}_{d}}, {\accVec{}{}}_{d}),
L_{\text{color}} = \sum_{d} \langle \left( \BRIGHT{}{d}, \accVec{}{{d}} \right),
\label{ang}
\end{equation}

\noindent where $d$ indicates the pixel position and $< ( , )$ denotes the calculation of the angular difference between two vectors in the three-dimensional RGB space.

\textbf{Brightness Loss:} Its functionality is based on the L1 loss function, but in this approach, it operates on grayscale images to ensure brightness consistency between the input image and the resulting image. Supervision is performed using the average pixel intensities, as presented in equation~\eqref{Brightness} \cite{yin2023cle}.

\begin{equation}
L_{brightness}= I_g|(\BRIGHT{}{})-I_g(\accVec{}{})|_{1},
\label{Brightness}
\end{equation}

\noindent where $I_g(.)$ represents the version in gray-scale of a RGB image.

\textbf{MSE Loss:} The Mean Squared Error (MSE) function measures how well a model performs by calculating the square of the distance (i.e., the error) between the predicted and actual values. In other words, the closer the predicted value is to the actual value, the lower the mean squared error between the two. Its equation is as follows:

\begin{equation}
L_{MSE}= \frac{1}{N}\sum_{d=1}^{N}({\BRIGHT{}{}}-{\accVec{}{}})^{2}.
\label{mse}
\end{equation}

\noindent where $N$ is the number of samples, $\accVec{}{}$ represents the actual value of the $d-th$ sample, $\BRIGHT{}{}$ represents the predicted value of the $d-th$ sample.

%=================================================================

\section{Experiments}
\label{sec:experiments}

This section presents a qualitative and a quantitative evaluation of the proposed approach, benchmarking it against traditional and state-of-the-art techniques in deep learning.

\subsection{Experimental Datasets} 

In the experiments, the challenging UIEB, SUIM and RUIE  underwater image datasets are used. The mentioned datasets are composed of underwater images from the sea and are widely used in the literature \cite{UIEB}\cite{SUIM}\cite{RUIE}. The UIEB underwater image dataset is composed of 890 images of natural environments, the SUIM underwater image dataset is composed of 1635 images of natural environments, while the RUIE underwater image dataset is composed of 4230 images of natural environments.

\subsection{Comparison and Metrics} 

Our brightness enhancement method is compared with well-established techniques, including RUIDL \cite{Martinho2024}, UESAM \cite{Claudio} and UDNet \cite{saleh2023distribuicao}. To evaluate and compare the results we use the metrics PSNR \cite{PSNR}, SSIM \cite{wang2004image}, UIQM \cite{UIQM}, and UISM \cite{UIQM}. We aim to comprehensively assess the effectiveness of our method in comparison to these literature techniques.

\subsection{Implementation Details} 

Our approach uses OpenCV and PyTorch frameworks on a Workstation, with an Intel$^{\text{\textregistered}}$ Core$^{TM}$ i7-10700K CPU @ 3.80GHz, 64 GB DDR4-3000 main memory and an NVIDIA$^{\text{\textregistered}}$ Titan$^{\text{\textregistered}}$ RTX 24 GB GDDR6.
For the training stage, we defined batch size equal to 8, 1000 epochs and learning rate equal to $5\times10^{-5}$. The AdamW optimization algorithm is used for training UDBE, with weight decay equal to $1\times10^{-4}$, regarding dataset split of 90\% for train and 10\% for test. 

For the learning process, the LPIPS, SSIM, and MSE loss functions were first used in training, addressing more general and structural features in underwater images. After the first 20 training epochs, the color and brightness loss functions were included in the training to refine the previous knowledge by considering more specific and relevant features for brightness enhancement of underwater images.

%The learning stage considers the loss functions LPIPS, SSIM, and MSE at the beginning of the training process, addressing the most general and structural features in underwater images. After the first 20 epochs, the training of the color and brightness loss functions starts, refining the previous knowledge in more specific and relevant features for the brightness enhancement of subaquatic images.

\subsection{Qualitative Evaluation} 

In the qualitative analysis, we compare the proposed underwater image enhancement technique with the literature methods. We assess the visual quality and perceptual enhancements across scenes from all datasets. Through the comparison of our technique's results with those from other methods, we intend to highlight differences in image brightness, visibility, and overall visual quality. This qualitative evaluation offers valuable insights into the strengths and limitations of our approach. Additionally, this evaluation method serves as a fundamental step in validating the quality of our proposed technique for brightness enhancement in underwater images.

\begin{figure*}[!ht]
    \centering
    \includegraphics[width=0.17\linewidth]{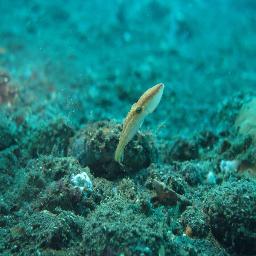}
    \hspace{-.17cm}
    \includegraphics[width=0.17\linewidth]{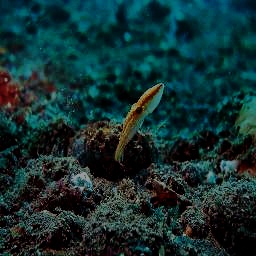}
    \hspace{-.17cm}
    \includegraphics[width=0.17\linewidth]{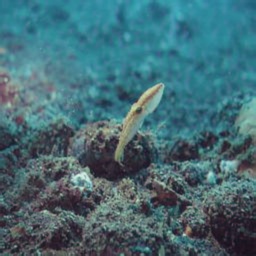}
    \hspace{-.17cm}
    \includegraphics[width=0.17\linewidth]{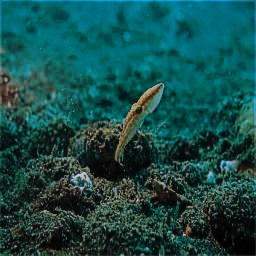}
    \hspace{-.17cm}
    \includegraphics[width=0.17\linewidth]{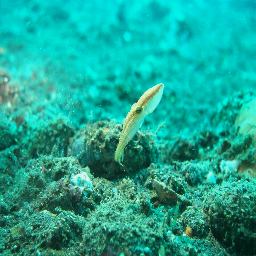}
    \hspace{-.17cm}
%    \includegraphics[width=0.14\linewidth]{Images/AUID2_1-RAW.png}
%    \hspace{-.17cm}
%    \includegraphics[width=0.14\linewidth]{Images/AUID2_1-RAW.png}
%    \hspace{-.17cm}
    \vspace{-0.35cm}
\end{figure*}
%\vspace{-5.25cm}
 \begin{figure*}[!ht]
     \centering
     \includegraphics[width=0.17\linewidth]{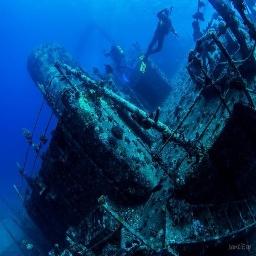}
     \hspace{-.17cm}
     \includegraphics[width=0.17\linewidth]{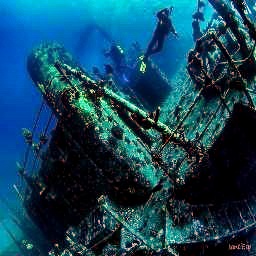}
     \hspace{-.17cm}
     \includegraphics[width=0.17\linewidth]{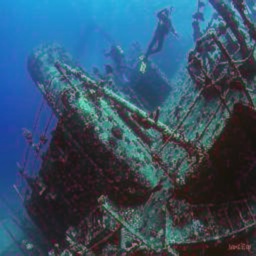}
     \hspace{-.17cm}
     \includegraphics[width=0.17\linewidth]{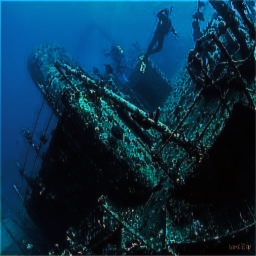}
     \hspace{-.17cm}
     \includegraphics[width=0.17\linewidth]{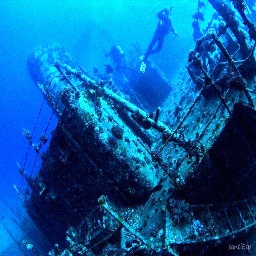}
     \hspace{-.17cm}
%     \includegraphics[width=0.14\linewidth]{Images/AUID2_1-RAW.png}
%     \hspace{-.17cm}
%     \includegraphics[width=0.14\linewidth]{Images/AUID2_1-RAW.png}
%     \hspace{-.17cm}
     \vspace{-0.75cm}
 \end{figure*}
%\vspace{-5.25cm}
\begin{figure*}[!ht]
 %    \captionsetup[subfloat]{labelfont=scriptsize,textfont=scriptsize}
    \centering
    \hspace{-.04cm}
    \subfloat[Raw]{\includegraphics[width=0.17\linewidth]{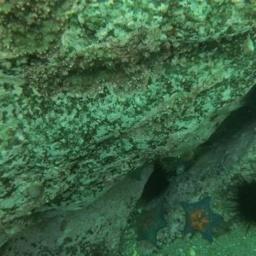}}
    \hspace{-.04cm}
    \subfloat[RUIDL]{\includegraphics[width=0.17\linewidth]{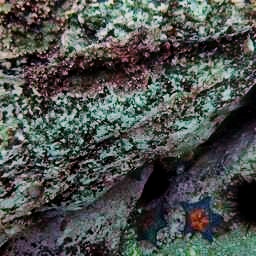}}
    \hspace{-.04cm}
    \subfloat[UDNet]{\includegraphics[width=0.17\linewidth]{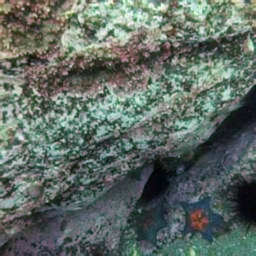}}
    \hspace{-.04cm}
    \subfloat[UESAM]{\includegraphics[width=0.17\linewidth]{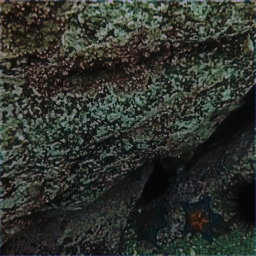}}
    \hspace{-.04cm}
    \subfloat[UDBE (Ours)]{\includegraphics[width=0.17\linewidth]{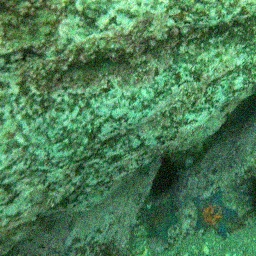}}
    \hspace{.16cm}
%    \subfloat[Medium UDBE]{\includegraphics[width=0.14\linewidth]{Images/AUID2_1-RAW.png}}
%    \hspace{-.07cm}
%    \subfloat[High UDBE]{\includegraphics[width=0.14\linewidth]{Images/AUID2_1-RAW.png}}
    %\hspace{-.07cm}
    
    \caption{Qualitative comparison of restored underwater images on the UIEB (1$^{st}$ row), SUIM (2$^{nd}$ row) and RUIE (3$^{rd}$ row) datasets. From left to right are presented the raw underwater images and the results of RUIDL, UDNet, UESAM, and our approach UDBE.}
    \label{fig:qualitative}

\end{figure*}

Figure \ref{fig:qualitative} presents the output underwater images after applying the underwater enhancement algorithms, taking into account the UIEB, SUIM and RUIE datasets. The first row presents the results regarding the UIEB dataset. The second row presents the results regarding the SUIM dataset. Meanwhile, the third row presents the results regarding the RUIE dataset. 
Additionally, are evaluated the techniques: RUIDL, UESAM, UDNet and UDBE (Ours). Through Figure \ref{fig:qualitative} it is possible to verify that our approach presents significant results in the brightness enhancement of underwater scenes in all datasets. In addition, it is possible to observe the dark outcomes obtained using UESAM method, while the UDNet algorithm is the comparison technique with more consistent results, although presents blurring effects in the enhanced images. The RUIDL presented significant enhancement in SUIM and RUIE datasets, however the proposed UDBE method achieved brightness improvement, good visibility and meaningfull visual quality, for all challenging datasets and considering relevant quality metrics.

%\felipe{Verificar como melhorar a discussão em função das 3 imagens do UDBE. Destacar pq mostrar as 3 imagens do UDBE. Verificar com o Gustavo}

\subsection{Quantitative Evaluation}

In the quantitative analysis, we assess the outcomes obtained by the proposed method (UDBE) and the comparison literature methods (UDnet, UESAM e RUIDL). In this evaluation, the achieved results are defined as quantitative measurements, indicating the accuracy of the enhancement process.

In order to improve the quantitative evaluation process, full reference and non-reference quality metrics are used. For full reference analysis, through PNSR and SSIM metrics, the generated high-brightness underwater images are considered as reference images, and the low-brightness underwater images as input images. For non-reference analysis, through UIQM and UISM metrics, low-brightness underwater images are considered as input images. It is worth mentioning that, for all quality metrics, the higher the metric value, the more accurate the brightness enhancement. For the PSNR metric, the results denote that the output image is closer to the reference image in terms of image content. For the SSIM metric, the result is more similar to the reference in terms of image structure and texture. For UIQM, the results involve the analysis of image colorfulness, sharpness, and contrast. Finally, the UISM metric evaluates the image sharpness.

In Table \ref{quantitative} we compare the proposed approach and the algorithms RUIDL, UESAM and UDNet, using full reference and non-reference image quality metrics. From the results in Table~\ref{quantitative}, it was possible to verify that our approach outperforms the comparison methods, even considering three challenging datasets and different quality metrics. Regarding all datasets, the UDNet and RUIDL enhancement techniques presented the best results among the comparison techniques. However, the results regarding different and challenging metrics and datasets demonstrate the robustness of the proposed approach in the evaluated scenarios.

%\felipe{Aguardar os resultados para dar mais detalhes na discussão.}

\begin{table}[!ht]
\centering
\caption{Full and Non-Reference Image Quality Assess-
ment in Terms of PSNR, SSIM, UIQM and UISM on UIEB, SUIM and RUIE underwater image datasets.}
\begin{tabular}{c|cccc}
\hline
\multirow{2}{*}{Methods} & \multicolumn{4}{c}{UIEB}                                                                                   \\ \cline{2-5} 
                         & \multicolumn{1}{l}{PSNR$\uparrow$} & \multicolumn{1}{l}{SSIM$\uparrow$} & \multicolumn{1}{l}{UIQM$\uparrow$} & \multicolumn{1}{l}{UISM$\uparrow$} \\ \hline
RUIDL \cite{Martinho2024}            & 11.49                       & 0.390                       & 0.789                       & 3.464                        \\
UDNet \cite{saleh2023distribuicao}             & 19.40                       & 0.611                      & 0.558                        & 4.004                         \\
UESAM \cite{Claudio}           & 10.18                       & 0.145                       & 0.559                       & 4.041                        \\
\textbf{UDBE (Ours)}     & \textbf{23.54}              & \textbf{0.8560}              & \textbf{0.9284}              & \textbf{5.219}               \\ \hline
\multirow{2}{*}{Methods} & \multicolumn{4}{c}{SUIM}                                                                                   \\ \cline{2-5} 
                         & \multicolumn{1}{l}{PSNR$\uparrow$} & \multicolumn{1}{l}{SSIM$\uparrow$} & \multicolumn{1}{l}{UIQM$\uparrow$} & \multicolumn{1}{l}{UISM$\uparrow$} \\ \hline
RUIDL \cite{Martinho2024}            & 17.33                       & 0.706                       & 0.788                       & 4.188                        \\
UDNet \cite{saleh2023distribuicao}             & 20.74                       & 0.738                       & 0.456                      & 3.422                       \\
UESAM \cite{Claudio}           & 10.13                       & 0.252                       & 0.317                       & 3.247                        \\
\textbf{UDBE (Ours)}     & \textbf{25.83}              & \textbf{0.8618}              & \textbf{0.9147}              & \textbf{4.624}               \\ \hline
\multirow{2}{*}{Methods} & \multicolumn{4}{c}{RUIE}                                                                                   \\ \cline{2-5} 
                        & \multicolumn{1}{l}{PSNR$\uparrow$} & \multicolumn{1}{l}{SSIM$\uparrow$} & \multicolumn{1}{l}{UIQM$\uparrow$} & \multicolumn{1}{l}{UISM$\uparrow$} \\ \hline
RUIDL \cite{Martinho2024}            & 16.13                       & 0.399                       & \textbf{0.829}                       & \textbf{4.674}                        \\
UDNet \cite{saleh2023distribuicao}             & 16.01                       & 0.355                       & 0.624                      & 4.005                        \\
UESAM \cite{Claudio}           & 12.87                       & 0.407                       & 0.194                      & 3.483                        \\
\textbf{UDBE (Ours)}     & \textbf{33.49}              & \textbf{0.9437}              & 0.4531              & 3.902               \\ \hline
\end{tabular}
\label{quantitative}
\end{table}
% UIEB
% psnr_orgin_avg:23.543456999095564
% ssim_orgin_avg:0.8560410680907343
% uiqm_orgin_avg:0.9283596696141649
% uciqe_orgin_avg:0.5636676565932603
% SUIM
% psnr_orgin_avg:25.834871459783212
% ssim_orgin_avg:0.861790453725576
% uiqm_orgin_avg:0.9147083231449237
% uciqe_orgin_avg:0.5767991469318321
% RUIE
% psnr_orgin_avg:33.492325286788876
% ssim_orgin_avg:0.9436814042088719
% uiqm_orgin_avg:0.45309976089750775
% uciqe_orgin_avg:0.5073949929822728

%=================================================================

\section{Conclusion}
\label{sec:conclusion}

In this paper we presented an approach for lighting enhancement in underwater images. For this, we proposed an unsupervised diffusion-based learning approach for mitigating the lighting degradation in subaquatic images.
From the obtained results we verify that the proposed method is accurate and presents clearer underwater images. Additionally, the experiments demonstrate the robustness of our approach, regarding different and challenging datasets and present quality results in distinct underwater scenarios, brightness levels and water colors. 
Finally, qualitative and quantitative assessment experiments demonstrated the effectiveness of our approach, which achieves reasonable performance in terms of brightness quality, even regarding different image quality metrics.
As future work, we intend to investigate other strategies to employ state-of-the-art diffusion models for the enhancement of underwater images, aiming to improve their quality and visibility. Additionally, we plan to evaluate the computational complexity of UDBE. We also intend to explore the relationship and impacts of loss functions in the underwater domain for image enhancement, providing solutions for subaquatic applications.

%=================================================================
\section*{Acknowledgment}

%The authors would like to thank the partner organizations for their research support and financial assistance. Furthermore, we are grateful to the creators of the UIEB, SUIM and RUIE datasets.

This study was financed, in part, by the São Paulo Research Foundation (FAPESP), Brasil. Process Number 2024/10523-5. The authors would also like to thank the PRH-ANP and CNPQ organizations for their research support and financial assistance.

\bibliographystyle{IEEEtran}
\bibliography{IEEEexample}

\begin{thebibliography}{10}
\providecommand{\url}[1]{#1}
\csname url@samestyle\endcsname
\providecommand{\newblock}{\relax}
\providecommand{\bibinfo}[2]{#2}
\providecommand{\BIBentrySTDinterwordspacing}{\spaceskip=0pt\relax}
\providecommand{\BIBentryALTinterwordstretchfactor}{4}
\providecommand{\BIBentryALTinterwordspacing}{\spaceskip=\fontdimen2\font plus
\BIBentryALTinterwordstretchfactor\fontdimen3\font minus \fontdimen4\font\relax}
\providecommand{\BIBforeignlanguage}[2]{{%
\expandafter\ifx\csname l@#1\endcsname\relax
\typeout{** WARNING: IEEEtran.bst: No hyphenation pattern has been}%
\typeout{** loaded for the language `#1'. Using the pattern for}%
\typeout{** the default language instead.}%
\else
\language=\csname l@#1\endcsname
\fi
#2}}
\providecommand{\BIBdecl}{\relax}
\BIBdecl

\bibitem{fayaz2020underwater}
S.~Fayaz, S.~Parah, G.~Qureshi, and V.~Kumar, ``Underwater image restoration: A state-of-the-art review,'' \emph{IET Image Processing}, vol.~15, 2020.

\bibitem{abbas2019}
S.~Abbas, ``Underwater image enhancement by fusion,'' Ph.D. dissertation, CAPITAL UNIVERSITY, 2019.

\bibitem{ho2020denoising}
J.~Ho, A.~Jain, and P.~Abbeel, ``Denoising diffusion probabilistic models,'' \emph{Advances in neural information processing systems}, vol.~33, pp. 6840--6851, 2020.

\bibitem{ronneberger2015u}
O.~Ronneberger, P.~Fischer, and T.~Brox, ``U-net: Convolutional networks for biomedical image segmentation,'' in \emph{Medical image computing and computer-assisted intervention--MICCAI 2015: 18th international conference, Munich, Germany, October 5-9, 2015, proceedings, part III 18}.\hskip 1em plus 0.5em minus 0.4em\relax Springer, 2015, pp. 234--241.

\bibitem{Martinho2024}
\BIBentryALTinterwordspacing
L.~A. Martinho, J.~M.~B. Calvalcanti, J.~L. Pio, and F.~G. Oliveira, ``Diving into clarity: Restoring underwater images using deep learning,'' \emph{Journal of Intelligent \& amp; Robotic Systems}, vol. 110, no.~1, Feb. 2024. [Online]. Available: \url{http://dx.doi.org/10.1007/s10846-024-02065-8}
\BIBentrySTDinterwordspacing

\bibitem{he2010single}
K.~He, J.~Sun, and X.~Tang, ``Single image haze removal using dark channel prior,'' \emph{IEEE transactions on pattern analysis and machine intelligence}, vol.~33, no.~12, pp. 2341--2353, 2010.

\bibitem{peng2017underwater}
Y.-T. Peng and P.~C. Cosman, ``Underwater image restoration based on image blurriness and light absorption,'' \emph{IEEE transactions on image processing}, vol.~26, no.~4, pp. 1579--1594, 2017.

\bibitem{peng2018generalization}
Y.-T. Peng, K.~Cao, and P.~C. Cosman, ``Generalization of the dark channel prior for single image restoration,'' \emph{IEEE Transactions on Image Processing}, vol.~27, no.~6, pp. 2856--2868, 2018.

\bibitem{huang2018shallow}
D.~Huang, Y.~Wang, W.~Song, J.~Sequeira, and S.~Mavromatis, ``Shallow-water image enhancement using relative global histogram stretching based on adaptive parameter acquisition,'' in \emph{MultiMedia Modeling: 24th International Conference, MMM 2018, Bangkok, Thailand, February 5-7, 2018, Proceedings, Part I 24}.\hskip 1em plus 0.5em minus 0.4em\relax Springer, 2018, pp. 453--465.

\bibitem{hassan2021retinex}
N.~Hassan, S.~Ullah, N.~Bhatti, H.~Mahmood, and M.~Zia, ``The retinex based improved underwater image enhancement,'' \emph{Multimedia Tools and Applications}, vol.~80, pp. 1839--1857, 2021.

\bibitem{zhuang2021bayesian}
P.~Zhuang, C.~Li, and J.~Wu, ``Bayesian retinex underwater image enhancement,'' \emph{Engineering Applications of Artificial Intelligence}, vol. 101, p. 104171, 2021.

\bibitem{chen2023enhancement}
E.~Chen, T.~Ye, Q.~Chen, B.~Huang, and Y.~Hu, ``Enhancement of underwater images with retinex transmission map and adaptive color correction,'' \emph{Applied Sciences}, vol.~13, no.~3, p. 1973, 2023.

\bibitem{UIEFIT}
L.~A. Martinho, F.~G. Oliveira, J.~M.~B. Cavalcanti, and J.~L.~S. Pio, ``Underwater image enhancement based on fusion of intensity transformation techniques,'' in \emph{2022 Latin American Robotics Symposium (LARS), 2022 Brazilian Symposium on Robotics (SBR), and 2022 Workshop on Robotics in Education (WRE)}, 2022, pp. 348--353.

\bibitem{wang2021uiec}
Y.~Wang, J.~Guo, H.~Gao, and H.~Yue, ``Uiec\^{} 2-net: Cnn-based underwater image enhancement using two color space,'' \emph{Signal Processing: Image Communication}, vol.~96, p. 116250, 2021.

\bibitem{li2021underwater}
C.~Li, S.~Anwar, J.~Hou, R.~Cong, C.~Guo, and W.~Ren, ``Underwater image enhancement via medium transmission-guided multi-color space embedding,'' \emph{IEEE Transactions on Image Processing}, vol.~30, pp. 4985--5000, 2021.

\bibitem{wang2020gan}
J.~Wang, P.~Li, J.~Deng, Y.~Du, J.~Zhuang, P.~Liang, and P.~Liu, ``Ca-gan: Class-condition attention gan for underwater image enhancement,'' \emph{IEEE access}, vol.~8, pp. 130\,719--130\,728, 2020.

\bibitem{fu2021underwater}
B.~Fu, L.~Wang, R.~Wang, S.~Fu, F.~Liu, and X.~Liu, ``Underwater image restoration and enhancement via residual two-fold attention networks.'' \emph{Int. J. Comput. Intell. Syst.}, vol.~14, no.~1, pp. 88--95, 2021.

\bibitem{saleh2023distribuicao}
\BIBentryALTinterwordspacing
A.~Saleh, M.~Sheaves, D.~Jerry, and M.~Rahimi~Azghadi, ``Adaptive uncertainty distribution in deep learning for unsupervised underwater image enhancement,'' \emph{SSRN Electronic Journal}, 2023. [Online]. Available: \url{https://ssrn.com/abstract=4362438}
\BIBentrySTDinterwordspacing

\bibitem{Claudio}
\BIBentryALTinterwordspacing
C.~D. Mello, B.~U. Moreira, P.~J.~D. de~Oliveira~Evald, P.~J.~L. Drews, and S.~S. da~Costa~Botelho, ``Underwater enhancement based on a self-learning strategy and attention mechanism for high-intensity regions,'' \emph{Computers \&amp; Graphics}, vol. 107, p. 264–276, Oct. 2022. [Online]. Available: \url{http://dx.doi.org/10.1016/j.cag.2022.08.003}
\BIBentrySTDinterwordspacing

\bibitem{zhao2023wavelet}
C.~Zhao, W.~Cai, C.~Dong, and C.~Hu, ``Wavelet-based fourier information interaction with frequency diffusion adjustment for underwater image restoration,'' \emph{arXiv preprint arXiv:2311.16845}, 2023.

\bibitem{du2023uiedp}
D.~Du, E.~Li, L.~Si, F.~Xu, J.~Niu, and F.~Sun, ``Uiedp: Underwater image enhancement with diffusion prior,'' \emph{arXiv preprint arXiv:2312.06240}, 2023.

\bibitem{lu2023underwater}
S.~Lu, F.~Guan, H.~Zhang, and H.~Lai, ``Underwater image enhancement method based on denoising diffusion probabilistic model,'' \emph{Journal of Visual Communication and Image Representation}, vol.~96, p. 103926, 2023.

\bibitem{tang2023underwater}
Y.~Tang, H.~Kawasaki, and T.~Iwaguchi, ``Underwater image enhancement by transformer-based diffusion model with non-uniform sampling for skip strategy,'' in \emph{Proceedings of the 31st ACM International Conference on Multimedia}, 2023, pp. 5419--5427.

\bibitem{yin2023cle}
Y.~Yin, D.~Xu, C.~Tan, P.~Liu, Y.~Zhao, and Y.~Wei, ``Cle diffusion: Controllable light enhancement diffusion model,'' in \emph{Proceedings of the 31st ACM International Conference on Multimedia}, 2023, pp. 8145--8156.

\bibitem{perez2018film}
E.~Perez, F.~Strub, H.~De~Vries, V.~Dumoulin, and A.~Courville, ``Film: Visual reasoning with a general conditioning layer,'' in \emph{Proceedings of the AAAI conference on artificial intelligence}, vol.~32, no.~1, 2018.

\bibitem{song2020denoising}
J.~Song, C.~Meng, and S.~Ermon, ``Denoising diffusion implicit models,'' \emph{arXiv preprint arXiv:2010.02502}, 2020.

\bibitem{zhang2018unreasonable}
R.~Zhang, P.~Isola, A.~A. Efros, E.~Shechtman, and O.~Wang, ``The unreasonable effectiveness of deep features as a perceptual metric,'' in \emph{Proceedings of the IEEE conference on computer vision and pattern recognition}, 2018, pp. 586--595.

\bibitem{wang2004image}
Z.~Wang, A.~C. Bovik, H.~R. Sheikh, and E.~P. Simoncelli, ``Image quality assessment: from error visibility to structural similarity,'' \emph{IEEE transactions on image processing}, vol.~13, no.~4, pp. 600--612, 2004.

\bibitem{wang2019underexposed}
R.~Wang, Q.~Zhang, C.-W. Fu, X.~Shen, W.-S. Zheng, and J.~Jia, ``Underexposed photo enhancement using deep illumination estimation,'' in \emph{Proceedings of the IEEE/CVF conference on computer vision and pattern recognition}, 2019, pp. 6849--6857.

\bibitem{UIEB}
C.~Li, C.~Guo, W.~Ren, R.~Cong, J.~Hou, S.~Kwong, and D.~Tao, ``An underwater image enhancement benchmark dataset and beyond,'' \emph{IEEE Transactions on Image Processing}, vol.~29, pp. 4376--4389, 2020.

\bibitem{SUIM}
M.~J. Islam, C.~Edge, Y.~Xiao, P.~Luo, M.~Mehtaz, C.~Morse, S.~S. Enan, and J.~Sattar, ``{Semantic Segmentation of Underwater Imagery: Dataset and Benchmark},'' in \emph{IEEE/RSJ International Conference on Intelligent Robots and Systems (IROS)}.\hskip 1em plus 0.5em minus 0.4em\relax IEEE/RSJ, 2020.

\bibitem{RUIE}
R.~Liu, X.~Fan, M.~Zhu, M.~Hou, and Z.~Luo, ``Real-world underwater enhancement: Challenges, benchmarks, and solutions under natural light,'' \emph{IEEE Transactions on Circuits and Systems for Video Technology}, vol.~30, no.~12, pp. 4861--4875, 2020.

\bibitem{PSNR}
U.~Sara, M.~Akter, and M.~S. Uddin, ``Image quality assessment through fsim, ssim, mse and psnr—a comparative study,'' \emph{Journal of Computer and Communications}, vol.~07, pp. 8--18, 01 2019.

\bibitem{UIQM}
K.~Panetta, C.~Gao, and S.~Agaian, ``Human-visual-system-inspired underwater image quality measures,'' \emph{IEEE Journal of Oceanic Engineering}, vol.~41, no.~3, pp. 541--551, 2016.

\end{thebibliography}

\end{document}